\documentclass[11pt,a4paper]{article}
\usepackage[hyperref]{eacl2021}
\usepackage{times}
\usepackage{latexsym}

\usepackage{microtype}

\usepackage{algorithm}
\usepackage{algpseudocode}
\usepackage{subcaption}
\usepackage{graphicx}

\usepackage{amsmath}

\DeclareMathOperator*{\argmin}{arg\,min}

\usepackage{url}
\usepackage{enumitem}

\aclfinalcopy 

\newcommand{\ourframework}[0]{Language Transmission Simulator}

\title{Co-evolution of language and agents in referential games}

\author{Gautier Dagan \\
  University of Amsterdam \\
  {\tt gautierdagan@gmail.com} \\\And
  Dieuwke Hupkes \\
  Facebook AI Research\\
  {\tt dieuwkehupkes@fb.com} \\\AND
  Elia Bruni \\
  University of Osnabr\"{u}ck\\
  {\tt elia.bruni@gmail.com} \\
}

\date{}

\begin{document}
\maketitle

\begin{abstract}
Referential games offer a grounded learning environment for neural agents which accounts for the fact that language is functionally used to communicate.
However, they do not take into account a second constraint considered to be fundamental for the shape of human language: that it must be learnable by new language learners. 

\citet{Cogswell2019} introduced \emph{cultural transmission} within referential games through a changing population of agents to constrain the emerging language to be learnable.
However, the resulting languages remain inherently biased by the agents' underlying capabilities. 

In this work, we introduce \ourframework~to model both cultural and architectural evolution in a population of agents.
As our core contribution, we empirically show that the optimal situation is to take into account also the learning biases of the language learners and thus let language and agents \textit{co-evolve}.
When we allow the agent population to evolve through architectural evolution, we achieve across the board improvements on all considered metrics and surpass the gains made with \emph{cultural transmission}.
These results stress the importance of studying the underlying agent architecture and pave the way to investigate the \textit{co-evolution} of language and agent in language emergence studies.

\end{abstract}

\section{Introduction}
Human languages show a remarkable degree of structure and complexity.
In the evolution of this complex structure, several different intertwined \emph{pressures} are assumed to have played a role.
The first of these pressures concerns the function of language: as language is to communicate, it should allow effective communication between proficient language users \citep[e.g.][]{Smith2012}. 
This pressure is strongly intertwined with the nature of the proficient user: what features of language allow effective communication depends
on the abilities of the user to use the language.

A second pressure on the shape of human language stems from the fact that language must be \emph{learnable}.
Unlike animal languages, which are taken to be mostly innate, human languages must be re-acquired by each 
individual \cite{Pinker1990, Hurford99}.
A language can only survive if it can successfully be transmitted to a next generation of learners.
In the field of language evolution, this transmission process is referred to as \emph{cultural transmission}, while the process of change that occurs as a consequence is called \emph{cultural evolution}.\footnote{The importance of cultural evolution for the emergence of structure is supported by a number of artificial language learning studies \citep[e.g.][]{Saldana2018} and computational studies using the Iterated Learning paradigm, in which agents learn a language by observing the output produced by another agent from the previous `generation' \citep[e.g.][]{kalish2007,kirby2008,Kirby2015}.}
Like the pressures arising from the function of language, the way that cultural evolution shapes language also depends on the language users: what is learnable depends on the inductive biases of the learner.

Computationally, the emergence of language can be studied through simulation with artificial agents and by investigating the resulting languages for structure, level of compositionality, and morphosyntactic properties \citep{Kirby2001, Kirby2002}.
Originally based on logic and symbolic representations \cite{Kirby2001, Christiansen2003}, with the advent of modern deep learning methods, there has been a renewed interest in simulating the emergence of language through neural network agents \citep[i.a.][]{Lazaridou2016,Havrylov2017}.
Such work typically involves the use of \emph{referential games} \citep{lewis1969convention}, in which two or more agents have to emerge a language to obtain a shared reward.

These studies are motivated by the first pressure:  language as a tool for effective communication. 
However, they fail to consider the second pressure: language must be learnable by new agents.
They also fail to study the impact of the learning biases of the artificial agents themselves, which underlies both pressures.
In a recent study, \citet{Cogswell2019} proposed a method to include cultural evolution in a language emergence game. 
Their approach is more naturally aligned with pressures in humans language evolution than single agent referential games \citep[see e.g.][]{Wray2007}, but fails to account for the fact that cultural evolution and the learning biases of the artificial agents are two sides of the same coin: what language is learnable depends on the learning biases of the learner.

In this paper, we will therefore integrate the three components described above -- communication, cultural evolution and learning biases --  and setup a framework in which their interaction can be studied.
This framework, which we refer to with the term \ourframework, consists of \emph{a referential game}, played by \emph{a changing population of agents} -- simulating cultural evolution -- which are subject to \emph{architectural evolution} -- simulating the learning biases of the learners and allowing them to \emph{co-evolve} with the language.

Our contributions are three-fold fold:
\vspace{-2mm}\begin{itemize}\setlength{\itemsep}{-1mm}
        \item We introduce the \textit{\ourframework}, that allows to model both cultural and architectural evolution in a population of agents;
        \item We collect a large number of tests from previous work and combine them into an extensive test suits for language emergence games;
        \item We demonstrate that emerging languages benefit from including cultural transmission \emph{as well as} architectural evolution, but the best results are achieved when languages and agents can co-evolve.
\end{itemize}

\section{Related Work}
\label{sec:related_work}

%

Much work has been done on the emergence of language in artificial agents and investigating its subsequent structure, compositionality and morphosyntax \cite{Kirby2001, Kirby2002}.
Originally, such work was based on logic and symbolic representations \cite{Kirby2001, Christiansen2003}, but with the advent of modern deep learning \cite{lecun2015deep}, there has been a renewed interest in simulating the emergence of language through neural network agents \citep[i.a.][]{Foerster2016,Kottur2017,Choi2018,Lazaridou2016,Havrylov2017,Mordatch2017}. In the exploration of language emergence, different training approaches and tasks have been proposed to encourage agents to learn and develop communication. 
In a typical setup, two players aim to develop a communication protocol in which one agent must communicate information it has access to (typically an image), while the other must guess it out of a line-up \citep{Evtimova2017,Lazaridou2016}.

\citet{Kottur2017} show that `natural' language does not arise naturally in these communication games and it has to be incentivised by imposing specific restrictions on games and agents.
\citet{Havrylov2017} first demonstrated that using straight-through estimators were more effective than reinforcement learning in a collaborative task, and that optimizing rewards can lead to structured protocols (i.e. strings of symbols) to be induced from scratch. 
\citet{Mordatch2017} find that syntactic structure emerges in the  stream  of  symbol  uttered  by  agents, where symbols and syntax can be mapped to specific meanings or instructions. 
\citet{Choi2018} use qualitative analysis, visualization and a zero-shot test, to show that a language with compositional properties can emerge from environmental pressures.

\citet{chaabouni-etal-2019-word} find that emerged languages, unlike human languages, do not naturally prefer non-redundant encodings. 
\citet{chaabouni2020compositionality} further find that while generalization capabilities can be found in the languages, compositionality itself does not arise from simple generalization pressures.
\cite{rodriguez-luna-etal-2020-internal} encourage desirable properties of human languages, such as compositionality, to emerge through well-crafted auxiliary pressures.
Finally, \citet{harding-graesser-etal-2019-emergent} demonstrate with experiments on populations of agents that language can evolve and simplify through the interactions of different communities.

\citet{Cogswell2019} build upon the emergent language research by introducing cultural transmission as a pressure in referential games. 
They use a pool of agents with a resetting mechanism and show that this further encourages the emerging language to display compositional properties and structure allowing it to generalize better.
Pairing agents with one another in a larger population setting introduces cultural evolution, but it is the pressure introduced by the partial resetting which forces remaining agents to emerge a language that is quickly learnable by a new agent. While \citet{Cogswell2019} is the most related work to ours, an important difference is that they focus on cultural evolution only, without taking into account the learning biases of the agents via modelling architectural evolution.



\section{Approach}
\label{sec:approach}
In this paper, we introduce architectural evolution in language emergence games and study the interaction between cultural and architectural evolution with a range of different metrics.
Below, we first give a definition of the referential game we consider (Subsection~\ref{sec:referential_game}).
We then briefly explain our \ourframework~(Section ~\ref{sec:ourframework} and how we model cultural and architectural evolution within it ~(Section \ref{sec:cuevo} and \ref{sec:archevo} respectively).

\subsection{Referential games}\label{sec:referential_game}
We study language emergence in a referential game inspired by the signalling games proposed by \citet{lewis1969convention}.
In this game, one agent (the \textit{sender}) observes an image and generates a discrete message.
The other agent (the \textit{receiver}) uses the message to select the right image from a set of images containing both the sender image and several distractor images.
Since the information shown to the sender agent is crucial to the receivers success, this setup urges the two agents to come up with a communication protocol that conveys the right information.

Formally, our \textit{referential} game is similar to \citet{Havrylov2017}: 
{\small \begin{enumerate}[itemsep=-1ex]
    \item The meaning space of the game consists of a collection $D$ of $K$ images $\{d_0,d_1, ..., d_K\}$, represented by $z$-dimensional feature vectors.
    \item In each round $i$ of the game, a target item $d_i$ is randomly sampled from $D$, along with a set $C$ of $n$ distractor items.
    \item The sender agent $s$ of the game, parametrised by a neural network, is given item $d_i$, and generates a discrete message $m_i$ from a vocabulary $V$.
The message is capped to a max message length of $L$.
    \item The receiver agent $r$, also parametrised by a neural network, receives message $m_i$ and uses it to identify $d_i$ in the union of $d_i$ and $C$.
\end{enumerate}
}
\noindent We use $z=512$, and $n=3$ and train agents with Gumbel-Softmax \cite{jang2017categorical} based on task-success.


%

\subsection{\ourframework}
\label{sec:ourframework}
\begin{figure}[h]
	\centering
	\includegraphics[width=\columnwidth]{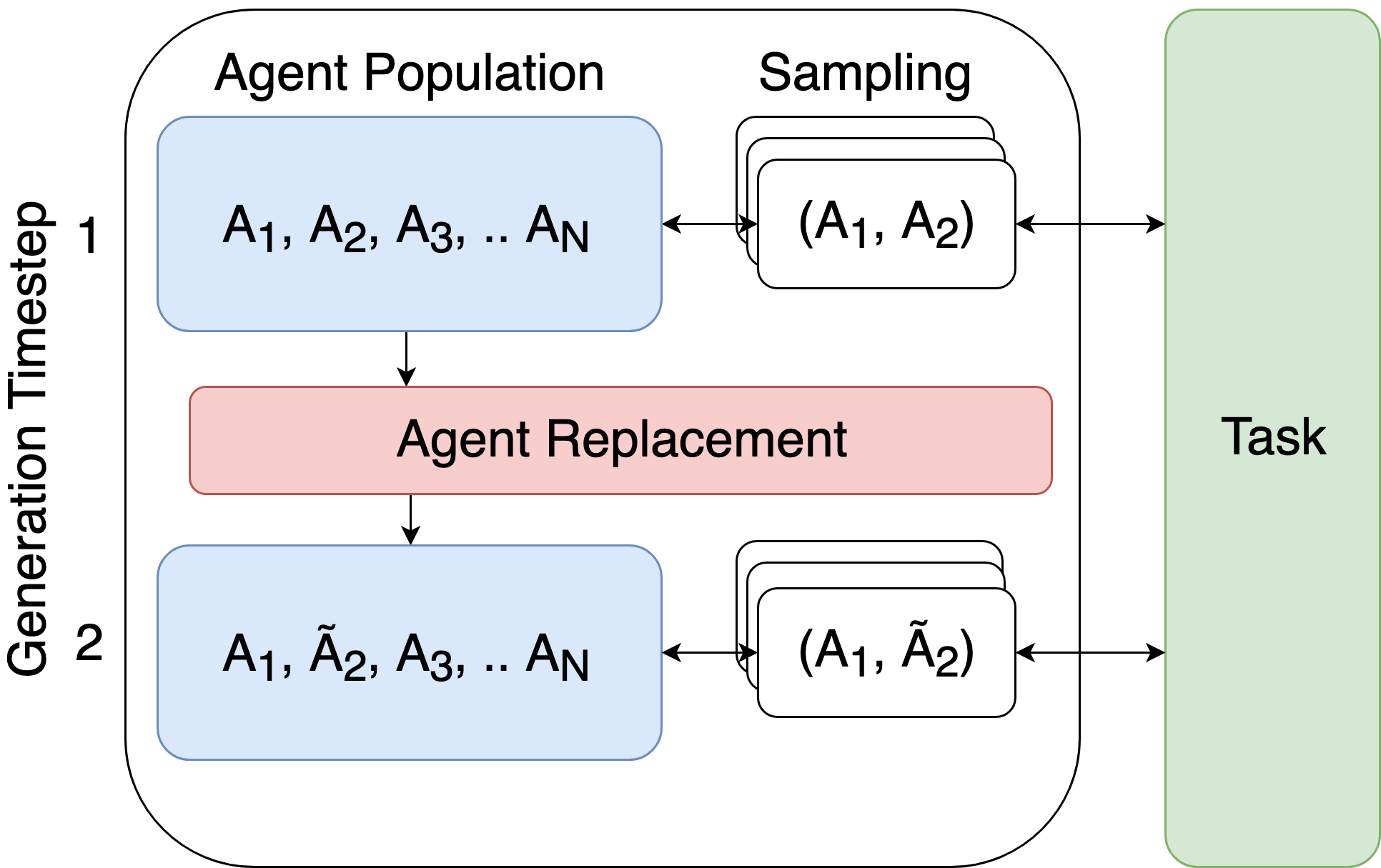}
	\caption{The \ourframework: Agent pairs are randomly sampled from each population and trained. After $l$ training steps, a portion $\alpha$ of the population is culled.}
	\label{fig:lte}
\end{figure}

In \ourframework, depicted in Figure~\ref{fig:lte}, we simulate a population of communicating agents.
In every training iteration, two random agents are sampled to play the game.
This forces the agents to adopt a simpler language naturally: to succeed they must be able to communicate or understand all opposing agents.
In our setup, agents are either sender or receiver, they do not switch roles during their lifetime.

\subsection{Cultural evolution in referential games}
\label{sec:cuevo}
Following \citet{Cogswell2019}, we simulate cultural evolution by periodically replacing agents in the population with newly initialised agents.
Cultural evolution is implicitly modelled in this setup, as new agents have to learn to communicate with agents that already master the task.
Following \citet{Cogswell2019}, we experiment with three different methods to select the agents that are replaced: randomly (no selection pressure), replacing the oldest agents or replacing the agents with the lowest fitness (as defined in Section~\ref{sec:fitness}).
We call these setups \texttt{cu-random}, \texttt{cu-age} and \texttt{cu-best}, respectively.

\subsection{Architectural evolution in referential games}
\label{sec:archevo}


To model architectural evolution, rather than periodically replacing agents with randomly initialised new agents, we instead mutate the most successful agents and replace the worst agents with variations of the best agents, as outlined in Section~\ref{sec:mutation}.
Note that cultural evolution is still implicitly modelled in this setup, as new agents still have to learn to communicate with older agents.
Therefore, we call this setup with the term \texttt{co-evolution}.

\subsubsection{Culling} We refer to the selection process and subsequent mutation or re-initialisation step as \textit{culling}. 
In biology, culling is the process of artificially removing organisms from a group to promote certain characteristics, so, in this case, culling consists of removing a subset of the worst agents and replacing them with variations of the best architecture.
The proportion of agents from each population selected to be mutated is determined by the culling rate $\alpha$, where $\alpha \in [0, 1)$.
The culling interval $l$ defines the number of iterations between culling steps.
A formalisation of the LTE can be found in appendix~\ref{app:lte}.

\subsubsection{Mutation Algorithm}\label{sec:mutation}
\label{sec:mutation_algo}

Our mutation algorithm is an intentionally simple implementation of a Neural Architectural Search (NAS).
NAS focuses on searching the architecture space of networks, unlike many traditional evolutionary techniques which often include parameter weights in their search space.
We opted to use the DARTS (Differentiable Architecture Search) RNN search space defined by  \citet{Liu2018}, which has obtained state-of-the-art performance on benchmark natural language tasks \cite{Li2019}.

The DARTS search space includes recurrent cells with up to $N$ nodes, where each node $n_1, n_2,...,n_N$ can take the output of any preceding nodes including $n_0$, which represents the cell's input.
All potential connections are modulated by an activation function, which can be the identity function, Tanh, Sigmoid or ReLU.
Following \citet{Liu2018} and \citet{Pham2018}, we enhance each operation with a highway bypass \cite{Zilly2016} and the average of all intermediate nodes is treated as the cell output.

To sample the initial model, we sample a random cell with a single node ($N=1$).
As this node must necessarily be connected to the input, the only variation stems from the possible activation functions applied to the output of $n_1$, resulting in four possible starting configurations.
We set a node cap of $N=8$.
We mutate cells by randomly sampling an architecture which is one edit step away from the previous architecture.
Edit steps are uniformly sampled from i) changing an incoming connection, ii) changing an output operation or iii) adding a new node; the mutation location is uniformly sample from all possible mutations.\footnote{For a formal description of the mutation process, we refer to Appendix~\ref{app:mutation}.}

\subsection{Fitness Criterion}
\label{sec:fitness}

The fitness criterion that we use in both the \texttt{cu-best} and \texttt{co-evolution} setup is based on task performance.
However, rather than considering agents' performance right before the culling step, we consider the age of the youngest agent in the population (defined in terms of number of batches that it was trained) and for every agent compute their performance up until \textit{when they had that age}.
For any agent $a_j$ in population $\mathbf{A}$ this is defined as: 
\begin{equation}
    \text{fitness}(a_j)= \frac{1}{\mathcal{T}_A}\sum_{t=0}^{\mathcal{T}_A} \mathcal{L}(a_j^t)
\end{equation}
where $\mathcal{T}_A = \min_{a \in A}\mathcal{T}(a)$ is the age $\mathcal{T}(a)$  of the youngest agent in the population, and $\mathcal{L}(a_j^t)$ is the loss of agent $a_j$ at time step $t$.
This fitness criterion is not biased towards older agents, that have seem already more data and have simply converged more.
It is thus not only considering task performance but also the \textit{speed} at which this performance is reached.






\section{Experiments}
\label{sec:experiments}


We test the LTE framework on a compositionally defined image dataset, using a range of different selection mechanisms.

 
\subsection{Dataset}
\label{subsec:dataset}
\begin{figure}[h]
	\centering
	\includegraphics[width=\columnwidth]{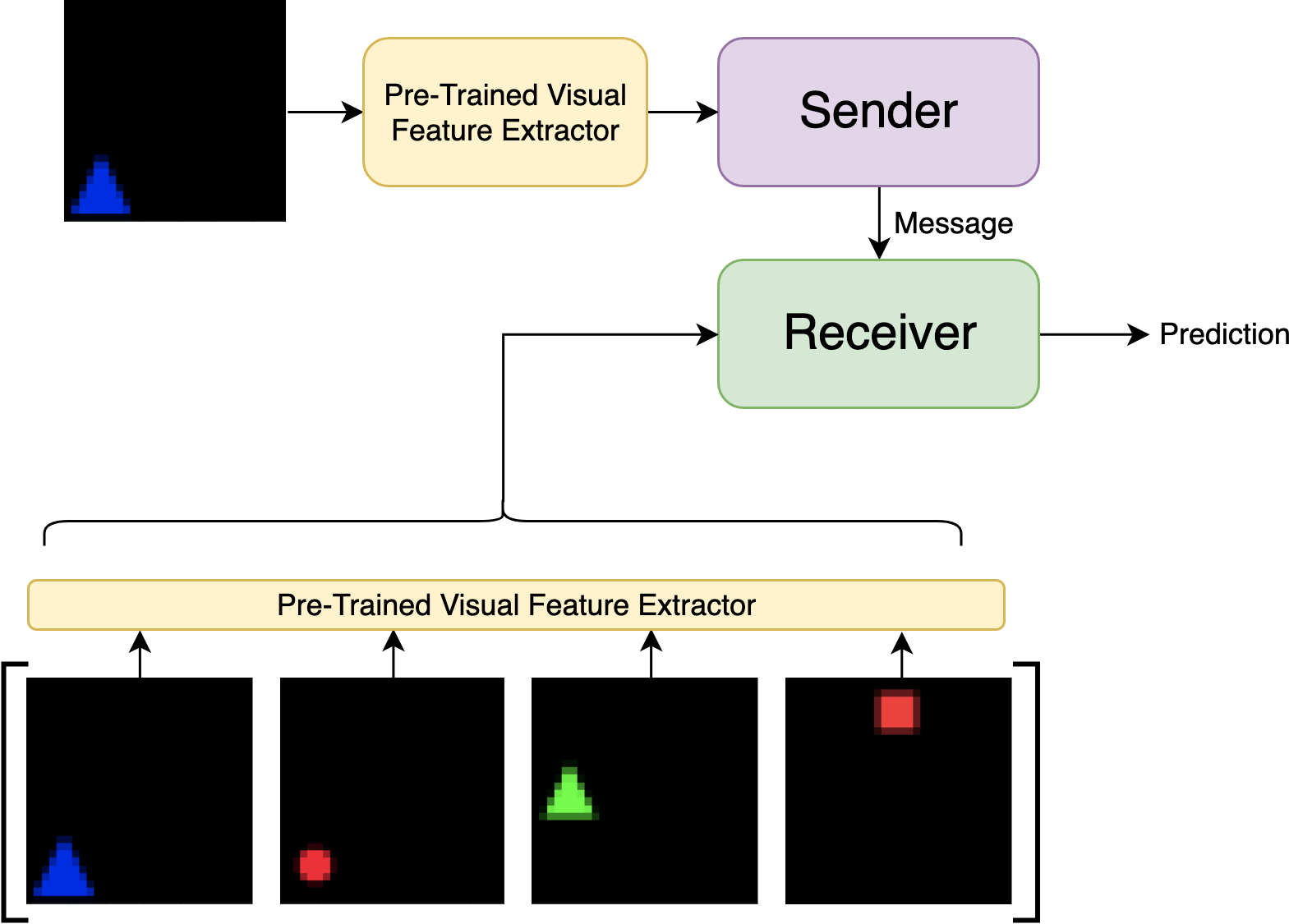}
	\caption{The modified Shapes task consists of showing an image the sender, and then letting the receiver deduce from the sender's message which image out of the target and $k$ distractors is the correct one.}
	\label{fig:shapes}
\end{figure}

In all our experiments, we use a modified version of the Shapes dataset \cite{Andreas2015}, which consists of 30 by 30 pixel images of 2D objects, characterised by shape (circle, square, triangle), colour (red, green, blue), and size (small, big). 
While every image has a unique symbolic description -- consisting of the shape, colour and size of the object and its horizontal and vertical position in a 3x3 grid -- one symbolic representation maps to multiple images, that differ in terms of exact pixels and object location.
We use 80k, 8k, 40k images for train, validation and test sets, respectively.
Some example images are depicted in Figure \ref{fig:shapes}. 

We pre-train a CNN feature extractor for the images in a two-agent setting of the task (see Appendix \ref{sec:features} for more details). 

\subsection{Architecture and Training}

For our co-evolution experiments, we use the DARTS search space as described above.
For all cultural evolution approaches, we use an LSTM \cite{lstm} for both the sender and receiver architecture (see Appendix \ref{sec:agent_architecture} for more details).
The sender and receiver models have a hidden size of 64 for the recurrent layer and an embedding layer of size 64. 
Further, we use a vocabulary size $V$ of 4, with an additional bound token serving as the indicator for beginning and end-of-sequence. 
We limit the maximum length of a sentence $L$ to 5. \footnote{The values for $V$ and $L$ were picked to provide a strong communication bottleneck to promote the emergence of structured and compressed languages, following the intuitions from \citet{Kottur2017} that natural language patterns do not emerge `naturally'.}

We back-propagate gradients through the discrete step outputs (message) of the sender by using the Straight-Through (ST) Gumbel-Softmax Estimator \citep{gumbel}.
We run all experiments with a fixed temperature $\tau=1.2$.
We use the default Pytorch \cite{paszke2017automatic} Adam \cite{adam} optimiser with a learning rate of 0.001 and a batch-size of 1024.
Note that the optimiser is reset for every batch.

For all multi-agent experiments we use a population size of 16 senders and 16 receivers. 
The culling rate $\alpha$ is set to $0.25$ or four agents, and we cull (re-initialise or mutate) every $l=5k$ iterations.
We run the experiments for a total of $I=500k$ iterations, and evaluate the populations before each culling step. 


\subsection{Evaluation}
 
We use an range of metrics to evaluate both the population of agents and the emerging languages.


\paragraph{Jaccard Similarity} We measure the consistency of the emerged languages throughout the population using \textit{Jaccard Similarity}, which is defined as the ratio between the size of the intersection and the union of two sets.
We sample $200$ messages per input image for each possible sender-receiver pair and average the Jaccard Similarity of the samples over the population. 
A high Jaccard Similarity between two messages is an indication that the same tokens are used in both messages. 

\paragraph{Proportion of Unique Matches}
We compute how similar the messages that different agents emit for the same inputs by looking at all possible (sender, message) pairs for one input and assess whether they are the same.
This metric is 1 when all agents always emit the same messages for the same inputs.

\paragraph{Number of Unique Messages}
We compute the average \textit{number of unique messages} generated by each sender in the population.
An intuitive reference point for this metric is the number of images with distinct symbolic representations.
If agents generate more messages than expected by this reference point, this demonstrates that they use multiple messages for the images that are -- from a task perspective -- identical.

\paragraph{Topographic Similarity} Topographic similarity, used in a similar context by \citet{lazaridou2018emergence}, represents the similarity between the meaning space (defined by the symbolic representations) and the signal space (the messages sent by an agent).
It is defined as the correlation between the distances between pairs in meaning space and the distances between the corresponding messages in the signal space.
We compute the topographic similarity for an agent by sampling 5,000 pairs of symbolic inputs and corresponding messages and compute the Pearson's $\rho$ correlation between the cosine similarity of the one-hot encoded symbolic input pairs and the cosine similarity of the one-hot encoded message pairs.


\paragraph{Average Population Convergence} To estimate the speed of learning of the agents in the population, estimate the \textit{average population convergence}.
For each agent, at each point in time, this is defined as the agents average performance from the time it was born until it had the age of the current youngest agent in the population (analogous to the fitness criterion defined in Section~\ref{sec:fitness}).
To get the average population convergence, we take we average those values for all agents in the population.

\paragraph{Average Agent Entropy} We compute the average certainty of sender agents in their generation process by computing and averaging their \textit{entropy} during generation.

\section{Results}
\label{sec:results}

We now present a detailed comparison of our cultural and co-evolution setups.
For each approach, we averaged over four random seeds, the error bars in all plots represent the standard deviation across these four runs.
To analyse the evolution of both agents and languages, we consider the development of all previously outlined metrics over time.
We then test the best converged languages and architectures in a single sender-receiver setup, to assess the impact of cultural and genetic evolution more independently.
In these experiments, we compare also directly to a single sender-receiver baseline, which is impossible for most of the metrics we consider in this paper.
Finally, we briefly consider the emerged architectures from a qualitative perspective.

\subsection{Task performance}

\begin{figure}[!h]
		\includegraphics[width=\columnwidth, trim=20mm 0 20mm 10mm, clip]{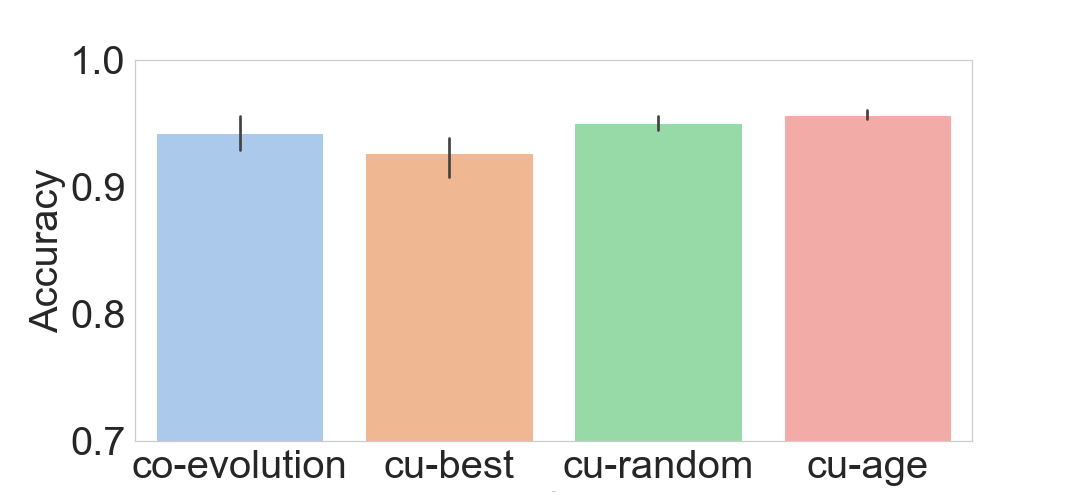}\hfill
	\centering
	\caption{Average Population Accuracy of final populations.}
	\label{fig:acc_loss}
\end{figure}

We first confirm that all setups in fact converge to a solution to the task.
As can be seen in Figure \ref{fig:acc_loss}, all populations converge to a (close to perfect) solution to the game. 
The \texttt{\normalsize cu-age} approach slightly outperforms the other approaches, with a accuracy that surpasses the $95\%$ accuracy mark.
Note that, due to the ever changing population, the accuracy at any point in time is an average of both `children' and `adults', that communicate with different members of the population. 



\section{Analysis}


In this section we analyse the resulting behaviour and success of agents in \ourframework.
We first use standard approaches such as average agent entropy and loss (convergence) to measure the success of agents with respect to their language and the task. 
Secondly, we use other metrics to analyse the emergent language itself in terms of consistency and diversity by using Jaccard Similarity, the proportion of unique matches, the number of unique messages, and the topographic similarity.
Thirdly, we perform a qualitative analysis of the architecture that emerge from our \ourframework.
Finally, we design Frozen Experiments, in which we test the emerged languages and architectures in a 1v1 setting with a fresh agent. 
This allows us to compare and measure the the improvement gains made by the architecture and those made by the language which emerged.
We show through these experiments that the co-evolution setting leads to a language that is both more successful and easier to learn for a given new agent. 

\subsection{Agent behaviour}

\begin{figure}[h!]
	\centering
	\includegraphics[width=\columnwidth, trim = 14mm 0mm 20mm 10mm]{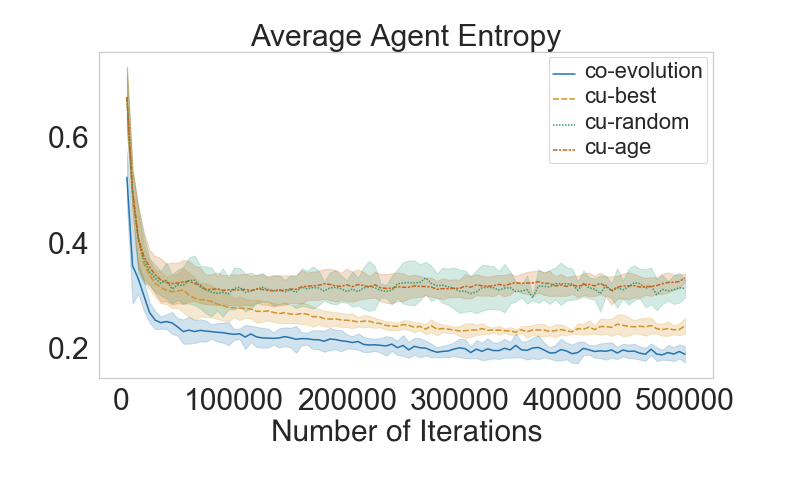}
	\caption{Average agent entropy over time.}
	\label{fig:ent}
\end{figure}

To assess the behaviour of the agents over time, we monitor their average message entropy convergence speed.
As can be seen in Figure~\ref{fig:ent}, the \texttt{\normalsize co-evolution} setup results in the lowest average entropy scores, the messages that they assign to one particular image will thus have lower variation than in the other setups.
Of the cultural evolution setups, the lowest entropy score is achieved in the \texttt{cu-best} setup.

\begin{figure}[h!]
	\centering
	\includegraphics[width=\columnwidth, trim= 14mm 0mm 20mm 10mm, clip]{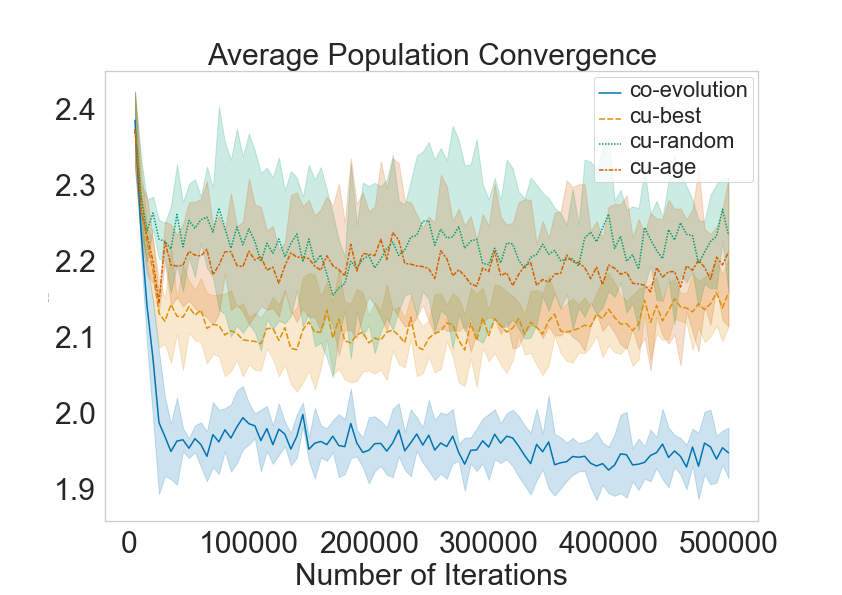}
	\caption{Average convergence for all cultural transmission modes and evolution.}
	\label{fig:convergence}
\end{figure}

Figure \ref{fig:convergence} shows the average population convergence over time. 
Also in this case, we observe a clear difference between cultural evolution only and co-evolution, with an immediately much lower convergence time for co-evolution and a slightly downward trending curve. 

\subsection{Language Analysis}

\begin{figure}[h]
	\includegraphics[width=\columnwidth, trim = 19mm 0 20mm 15mm, clip]{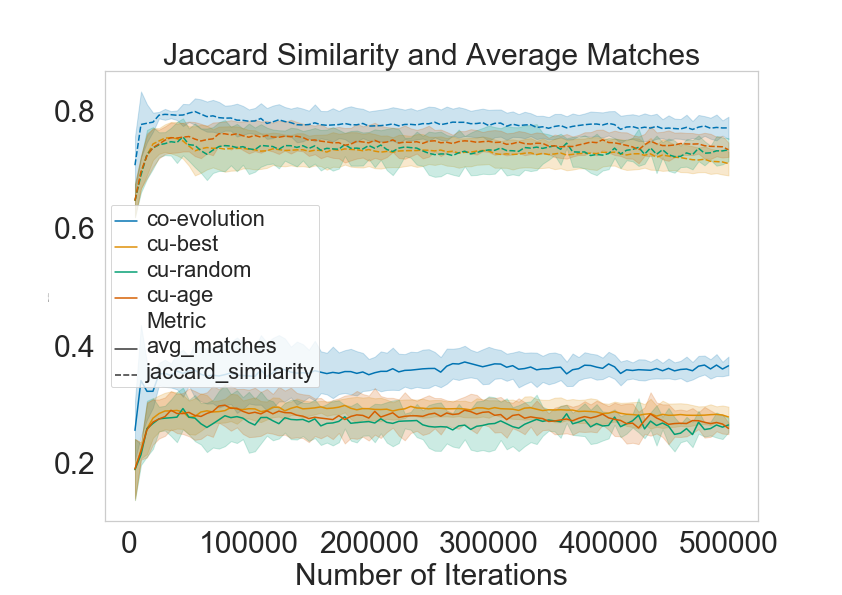}
	\caption{Average Jaccard Similarity and proportion of message matches for all cultural transmission modes and evolution}
	\label{fig:similarity}
\end{figure}

To check the consistencies of languages within a population, we compare the Jaccard Similarity and the Average Proportion of Unique Matches, which we plot in Figure \ref{fig:similarity}. This shows that, compared to cultural evolution only, not only are the messages in co-evolution more similar across agents (higher Jaccard Similarity), but also that agents are considerably more aligned with respect to the same inputs (less unique matches). 

\begin{figure}[h]
	\includegraphics[width=\columnwidth, trim = 13mm 0 2mm 21mm, clip]{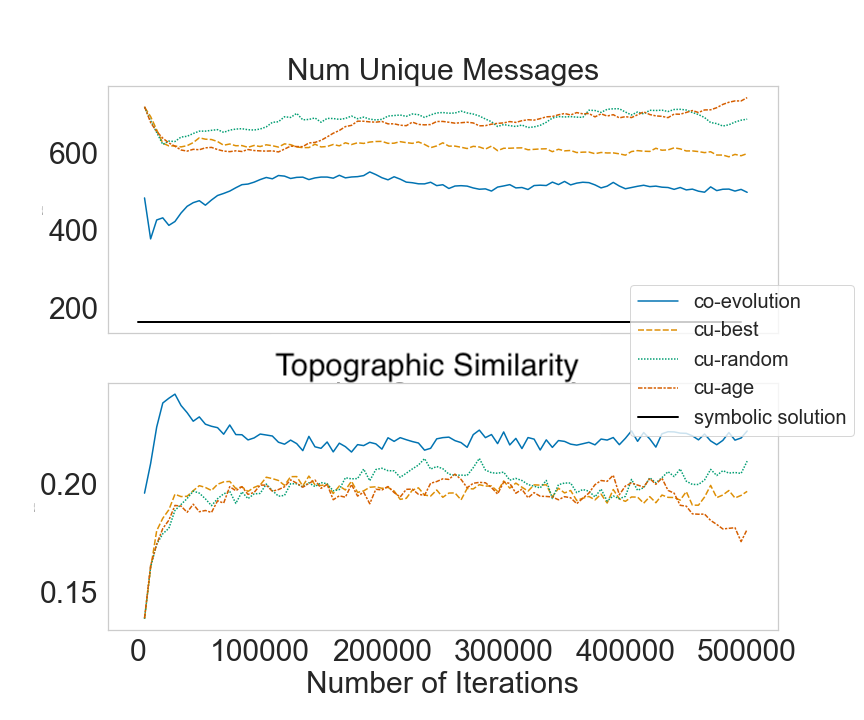}
	\caption{Average Number of Unique Messages and Topographic Similarity for all cultural evolution modes and co-evolution.
        For comparison, we also plot the number of unique messages for a symbolic solution that fully encodes all relevant features of the image (since we have three possible shapes and colours, two possible sizes, and a $3 \times 3$ grid of possible positions, this symbolic reference solution has $3 \times 3 \times 2 \times 9 = 162$ distinct messages.}
	\label{fig:topo}
\end{figure}


To assess the level of structure of the emerged languages, 
we plot the average Topographic Similarity and the Average Number of Unique Messages generated by all senders (Figure \ref{fig:topo}). 
The co-evolution condition again outperforms all cultural only conditions, with a simpler language (the number of the unique messages closer to the symbolic reference point) that is structurally more similar to the symbolic representation of the input (higher Topographical Similarity). 

\subsection{Architecture Analysis}
\label{sec:arch_analysis}

\begin{figure*}[h!]
	\centering
	\includegraphics[width=\textwidth]{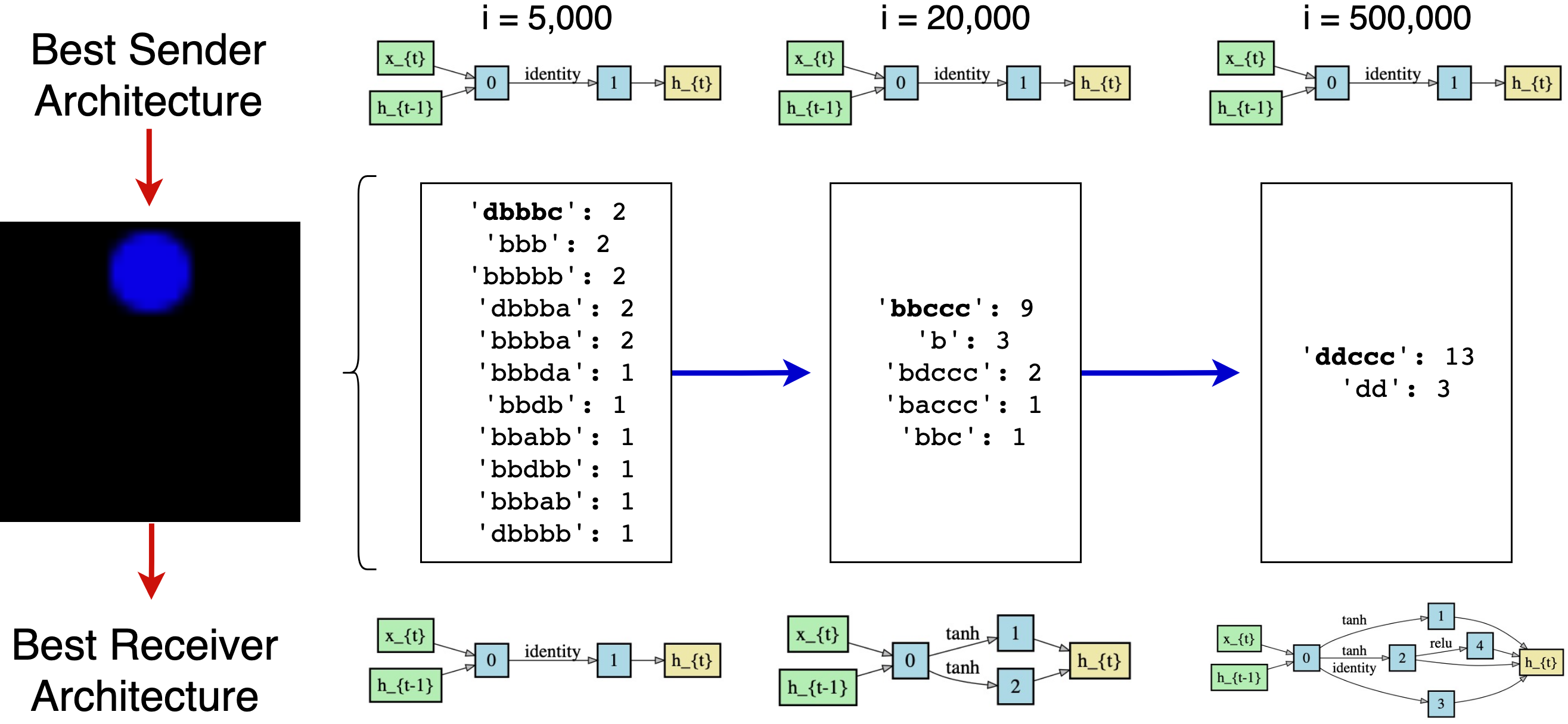}
	\caption{Evolution of the best sender and receiver architecture according to convergence, and the evolution of the population's message description of the same input through iterations. 
	The bold messages represent the message outputted by the best sender whose architecture is pictured above. 
	The count of each message represents the number of agents in the population which uttered this exact sequence. }
	\label{fig:qualitative}
\end{figure*}

In Figure \ref{fig:qualitative} we show the co-evolution of an agent and a sample of its language during three selected iterations in the co-evolution setup.
Strikingly, the best sender architecture does not evolve from its original form, which could point towards the limitations of of our search strategy and space.
On the contrary, the receiver goes through quite some evolution steps and converges into a significantly more complex architecture than its original form.
We observe a unification of language throughout evolution in Figure \ref{fig:qualitative}, which is also supported by Figure \ref{fig:topo}. 
The population of senders starts out 11 different unique messages and ends with only two to describe the same input image. 
We will leave more detailed analysis of the evolved architectures for future work.

\subsection{Frozen Experiments}

With a series of experiments we test the a priori suitability of the evolved languages and agents for the task at hand, by monitoring the accuracy of new agents that are paired with converged agents and train them from scratch.

We focus, in particular, on training receivers with a frozen sender from different setups, which allows us to assess 1) whether cultural evolution made languages evolve to be more easily picked up by new agents 2) whether the genetic evolution made architectures converge more quickly when faced with this task.
We compare the accuracy development of:\begin{itemize}
    \item An LSTM receiver trained with a frozen sender taken from \texttt{cu-best};
    \item An evolved receiver trained with a frozen evolved sender.
\end{itemize}
For both these experiments, we compare with two baselines:\begin{itemize}
    \item The performance of a receiver agent trained from scratch along with a receiver agent that has either the \texttt{cu} architecture or the evolved \texttt{co} architecture (\texttt{cu-baseline} and \texttt{co-baseline}, respectively);
    \item The performance of an agent trained with an agent that is \textit{pretrained} in the single agent setup, with either the \texttt{cu} architecture or an evolved architecture (\texttt{\normalsize cu-baseline-pretrained} and \texttt{co-baseline-pretrained}).
\end{itemize}

Each experiment is run 10 times, keeping the same frozen agent.
The results confirm cultural evolution contributes to the learnability and suitability of emerging languages: the \texttt{cu-best} accuracy (green line) converges substantially quicker and is substantially higher than the \texttt{cu-baseline-pretrained} accuracy (orange line). 
Selective pressure on the language appears to be important: the resulting languages are only easier to learn in the \texttt{cu-best} setup.\footnote{\texttt{cu-age} and \texttt{cu-random} are ommitted from the plot for clarity reasons.}
In addition, they show that the agents benefit also from the genetic evolution: the best accuracies are achieved in the co-evolution setup (red line).
The difference between the \texttt{cu-baseline} (blue) and the \texttt{co-baseline} (brown) further shows that even if the evolved architectures are trained from scratch, they perform much better than a baseline model trained from scratch.
The difference between the \texttt{co-baseline-pretrained} (only genetic evolution, purple line) and the co-evolution of agents and language line (red line) illustrates that genetic evolution alone is not enough: while a new evolved receiver certainly benefits from learning from a (from scratch) pretrained evolved sender, without the cultural transmission pressure, it's performance is still substantially below a receiver that learns from an evolved sender whose language was evolved as well.

\begin{figure}
	\includegraphics[width=\columnwidth, trim = 12mm 0 13mm 15mm, clip]{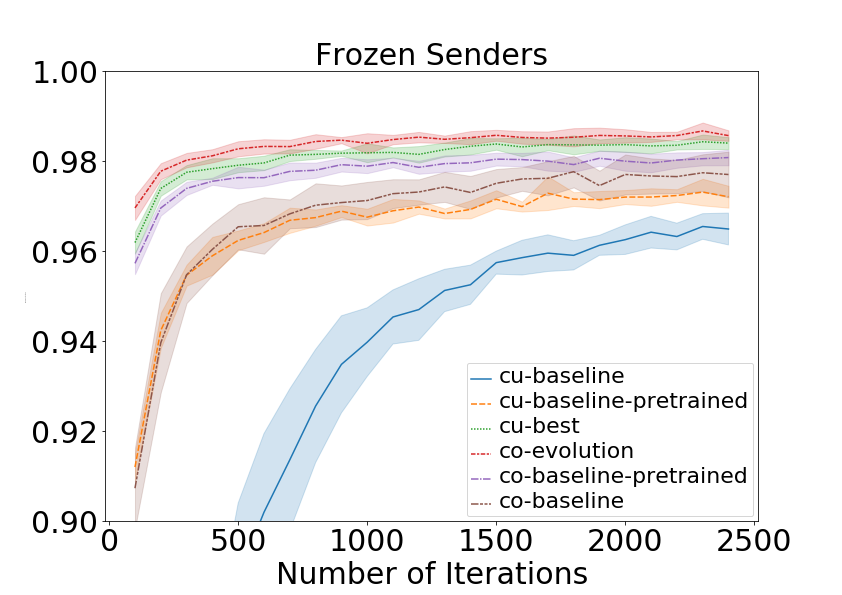}
	\caption{Receiver accuracies trained with different types of frozen senders.}
	\label{fig:frozen_sender}
\end{figure}

\section{Conclusion}
\label{sec:conclusion}
In this paper, we introduced a language transmission bottleneck in a referential game, where new agents have to learn the language by playing with more experienced agents. 
To overcome such bottleneck, we enabled both the cultural evolution of language and the architectural evolution of agents, using a new \ourframework. 
Using a battery of metrics, we monitored their respective impact on communication efficiency, degree of linguistic structure and intra-population language homogeneity. While we could find important differences in between cultural evolution strategies, it is when we included architectural evolution that agents scored best.
In a second experiment, we paired new agents with evolved languages and agents and again confirmed that, while cultural evolution makes a language easier to learn, co-evolution leads to the best communication.  

In future research, we would like to apply the \ourframework~on new, more complex tasks and further increase our understanding of the properties of the emerged languages and architectures.
Recent research has also found that relaxing the vocabulary size $V$ and sequence length $L$ constraints can lead to greater syntactic structure in emergent languages \cite{van-der-wal-etal-2020-grammar}.
We thus hope to investigate further relaxation of hyper-parameters and other neuro-evolution techniques in future work.


\section{Acknowledgements}
\label{sec:acknowledgements}
We would like to thank Angeliki Lazaridou for her helpful discussions and feedback on previous iterations of this work.


\bibliographystyle{acl_natbib}
\bibliography{main}

\appendix
\section{Appendix}
\label{sec:appendix}

\subsection{Language Transmission Engine}\label{app:lte}

We formalise our Language Transmission process in the pseudo code shown in Algorithm \ref{alg:lte}.
We select hyper-parameters $l$ as the number of iterations or batches shown between culling steps, and $I$ as the total number of iterations. 

\begin{algorithm}
	\caption{Language Transmission Engine}\label{alg:lte}
	\begin{algorithmic}
		\State $S \gets \{s_0, s_1 ... ,s_N \}$ 
		\State $R \gets \{r_0, r_1 ... ,r_N \}$ 
		\State $i \gets 1$
		\While {$i \leq I$}
		\For{batch $b$ in $D$}
		\State {Sample $\hat{s}$ from $S$}
		\State {Sample $\hat{r}$ from $R$}
		\State $\mathbf{train}(\hat{s}, \hat{r}, b)$
		\If {$i \bmod l = 0 $}
		\State {$\mathbf{cull}(S, R)$}
		\EndIf
		\State {$i \gets i+1$}
		\EndFor
		\EndWhile
	\end{algorithmic}

\end{algorithm}

\subsection{Mutation Algorithms Pseudo-code}\label{app:mutation}
\label{sec:mutation_pseudo_code}
\begin{algorithm}
\caption{Genotype-level Mutation}\label{alg:mutate_geno}
\begin{algorithmic}
\Procedure{$\mathbf{mg}$}{$genotype$}
	\State $g \gets copy(genotype)$
	\State {$a \gets \mathcal{U}(1, 3)$}
	\State {$n \gets \mathcal{U}(1, len(g))$}
	\If {$a = 1$}
		\State {$p \gets \mathcal{U}[ReLU, I, \tanh, \sigma]$} 
		\State {$n.activation \gets p$} 
	\EndIf
	\If {$a = 2$}
		\State  {$r \gets  \mathcal{U}(1, n)$} 
		\State {$n .connection \gets r$} 
	\EndIf
	\If {$a = 3$}
		\State {$n^\prime \gets new\_node()$}
		\State {$p \gets \mathcal{U}[ReLU, I, \tanh, \sigma]$} 
		\State  {$r \gets  \mathcal{U}(1, len(g))$} 
		\State {$n^\prime.activation \gets p$} 
		\State {$n^\prime.connection \gets r$} 
		\State {$g.append(n^\prime)$} 
	\EndIf
	\State \Return {$g$}
\EndProcedure

\end{algorithmic}
\end{algorithm}

The genotype mutation is described in pseudo-code by algorithm \ref{alg:mutate_geno}, and takes as input a genotype containing nodes describing the cell.
The genotype is mutated by either changing the input connection or primitive (output activation function) for a randomly sampled node $n$, or adding a new node altogether.
See section \ref{sec:mutation_algo} for explanations on the workings of the DARTS cell structure.

\begin{algorithm}
	\caption{Population-level Mutation}\label{alg:mutate_pop}
	\begin{algorithmic}
		\Procedure{$\mathbf{mutate}$}{$\mathbf{P}$}
			\State {$p^\prime \gets \argmin_{convergence}(P)$}
			\State {$\mathbf{p} \gets \pi(P)$}
			\For {$p_i$ in $\mathbf{p}$}
				\State {$p_i.genotype \gets \mathbf{mg}(p^\prime.genotype)$} 
			\EndFor
		\EndProcedure
	\end{algorithmic}
\end{algorithm}

In order to mutate a population $\mathbf{P}$ using $\pi$ as a replacement policy, we use the process outlined in algorithm \ref{alg:mutate_pop}.

\subsection{Agent Architecture}
\label{sec:agent_architecture}

\subsubsection{Sender Architecture}
The sender architecture comprises of a linear layer input mapping the input feature size (512) to the hidden size. 
The image feature vector is therefore mapped to the same dimension as the RNN layer, where it is used as the initial hidden state. 
When training, for each step of the sender RNN we apply the cell and use the straight-through Gumbel-Softmax trick to be able back-propagate gradients through the discrete message output.
During evaluation however, we sample the categorical distribution at each step to produce each token in the sentence.

\subsubsection{Receiver Architecture}
The receiver architecture is simpler and  takes as an input the message outputted by the sender and outputs a vector of input feature size (512).
A single embedding matrix is used to encode the sender's message.
During training the message is linearly transformed using the embedding matrix, while during the evaluation pass the discrete message outputs of the sender are used to map to the specific embedding dimensions.
The embedded message is then passed to the RNN layer, and the final state of the RNN is linearly mapped back to the feature size.
Doing so allows us to obtain a prediction for each image feature (distractors and true image), by comparing the alignment between the receiver output and the respective feature vectors.

\subsection{Feature Extraction}
\label{sec:features}

In order to obtain image features, we pre-trained a convolutional model on the task using the raw image as input.
Due to the input size requirements of the convolutional model, we resize the images linearly to be 128 by 128 (height, width) by 3 (RGB channels).
We used early stopping conditions on the validation accuracy, an embedding size of 256, and hidden size of 512.
The two agents are otherwise trained with the same parameters as other experiments: vocab size and max sentence length of 5, Adam optimizer with learning rate of 0.001.

For the visual module itself, we used a similar architecture to that in  \citet{Choi2018} albeit smaller. 
We used a five-layer convolution network with 20 filters, and a kernel size and stride of 3 for all layers.
For every convolutional layer, ReLU activation was applied on the output, after a Batch normalization step with no bias parameter.
The linear layer which followed the convolutional layers had output dimensions of 512 and a ReLU activation function. 
This allows us to obtain image features of size 512, which we then used for all experiments.

\subsection{Additional Figures and Analysis }
\label{sec:appendix_analysis}

\begin{figure}[h]
	\includegraphics[width=\columnwidth, trim = 20mm 0mm 20mm 0mm]{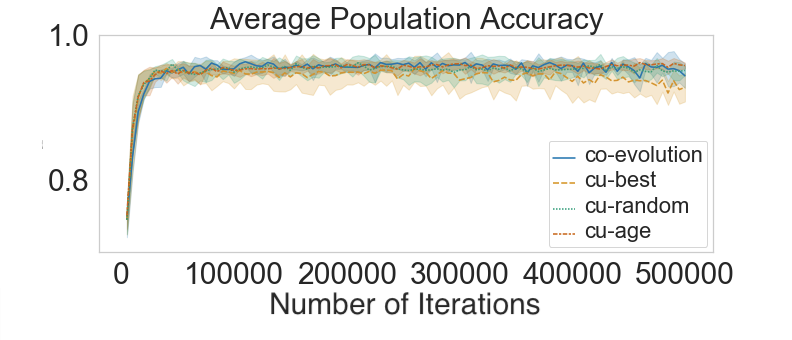}
	\caption{Average Population Accuracy for all Iterations}
	\label{fig:acc_iter}
\end{figure}

\end{document}